\documentclass[11pt,a4paper]{article}

\usepackage[hmargin=1.0in, vmargin=0.9in]{geometry}
\usepackage[table]{xcolor}  

\usepackage{graphicx}
\usepackage{caption,subcaption}
\usepackage{mathtools}
\usepackage{amsfonts}
\usepackage[inline]{enumitem}  
\usepackage{hyperref}
\hypersetup{
colorlinks=true,
citecolor=red
}

\usepackage{multicol}
\usepackage{tikz}
\usepackage{smartdiagram}
\usepackage[round]{natbib}  

\graphicspath{{fig/}}  

\usepackage[framemethod=tikz,innerleftmargin=3pt,innerrightmargin=3pt,leftmargin=-3pt,rightmargin=-3pt,hidealllines=true]{mdframed}

\begin{document}\sloppy

\title{A machine learning approach for efficient uncertainty \\
  quantification using multiscale methods\footnotetext{Submitted to Journal of Computational Physics (2017): \url{https://doi.org/10.1016/j.jcp.2017.10.034}}}

\author{Chan, Shing \and Elsheikh, Ahmed H.}

\date{}





\maketitle

\begin{abstract}

  Several multiscale methods account for sub-grid scale features using
  coarse scale basis functions. For example, in the Multiscale Finite
  Volume method the coarse scale basis functions are obtained by
  solving a set of local problems over dual-grid cells. We introduce a
  data-driven approach for the estimation of these coarse scale basis
  functions. Specifically, we employ a neural network predictor
  fitted using a set of solution samples from which it learns to
  generate subsequent basis functions at a lower computational cost
  than solving the local problems. The computational advantage of this
  approach is realized for uncertainty quantification tasks where a
  large number of realizations has to be evaluated. We attribute the
  ability to learn these basis functions to the modularity of the
  local problems and the redundancy of the permeability patches
  between samples. The proposed method is evaluated on elliptic
  problems yielding very promising results.

\end{abstract}


\section{Introduction}
Uncertainty quantification is an important task in practical
engineering where some parameters are unknown or highly
uncertain~\citep{Elsheikh2012,Elsheikh2013-wrr,Elsheikh2015_adv}. After
selecting adequate priors for the uncertain parameters, simulations
are performed for a large number of realizations. In the particular
case of reservoir simulations, the problem is further aggravated where
very fine details of the geological models are needed (large number of
cells) for accurate description of the flow.  One traditional approach
to address this problem is to upscale the geological
models~\cite{Babaei2013}. Another more recent approach is to use
multiscale methods~\citep{jenny2003multi,hou1997multiscale}. In these
methods, the global fine scale problem is decomposed into many smaller
local problems. The solution of these smaller local problems results
in a set of numerically computed basis functions which are then used
to build a coarse system of equations. After solving the coarse
system, an interpolation is performed with the basis functions to
obtain the fine scale solution.

We note that in multiscale methods, a large number of local problems
are solved under the same boundary conditions to obtain the required
basis functions. This process is repeated for each geological
realization in uncertainty quantification tasks. Our aim is to exploit
the redundancy that may arise in solving these local problems for
several geological realizations by introducing a data-driven approach
for estimating the basis functions efficiently. Specifically, we
exploit the large number of local problem solutions that become
available after a given number of full runs to construct a
computationally cheap function that generates approximate solutions to
local problems, i.e. approximate basis functions. In effect, what we
propose is a type of hybrid surrogate model by embedding a data-driven
approach into the multiscale numerical method. In this work, we focus
on one multiscale method called the Multiscale Finite Volume method
(MsFV) introduced by \citet{jenny2003multi}. However, the proposed
approach can be applied to any multiscale method where the explicit
construction of basis functions is performed such as in the more
recent multiscale method based on restriction-smoothed basis functions
(MsRSB) \citep{moyner2015multiscale}.

  \citet{aarnes2008mixed} introduced a multiscale mixed finite element
(MsMFE) method for porous media flows with stochastic permeability
field where a set of precomputed basis functions is constructed based
on selected set of realizations of the permeability field.  These
basis functions are then used to build a low-dimensional approximation
space for the velocity field. We note that the cost of solving the
upscaled problem (i.e. coarse scale) in \citep{aarnes2008mixed}
increases with the number of selected set of realizations.  In
contrast, in this manuscript we directly address the generation of
basis functions via a ``black box'' surrogate modeling approach using
machine learning. The generated basis functions are then directly
employed in the multiscale formulation without any further
modification. Another major difference is that our method directly
benefits from increasing the number of realizations used to build the
surrogate model without any increase in the computational cost of
solving the coarse scale problems for new realizations.

This paper presents the first attempt to combine/embed machine
learning techniques within multiscale numerical methods with very
promising results. The motivation for our work comes from the
observation that computational power and data storage capacity are
ever increasing. This trend is likely to continue for some time and
results in an increased ability to store and data-mine large volumes of
simulation data. Another source of motivation comes from the renewed
interest in machine learning among the research community, specially
in the branch of deep learning to tackle AI-complete tasks such as
computer vision and natural language processing. Neural network models
are regarded as \emph{universal function approximators}
\citep{hornik1989multilayer, hornik1990universal,
  cybenko1989approximation} with capacity to learn highly non-linear
maps. Therefore, they seem to be suitable for our current application.

The rest of the paper is organized as follows: In section 2, we give a
brief description of the multiscale finite volume method (MsFV) and
neural networks (NN). In section 3, we present the methodology for the
proposed approach for machine learning the basis. In section 4, we
examine the effectiveness of the presented method for uncertainty
quantification in two test cases. Finally, in section 5 we report the
conclusions of this work along with a brief discussion of future
directions.

\section{Background} 
In this section we briefly describe the two main components of the
proposed method: multiscale finite volume (MsFV) methods and neural
networks (NN) for surrogate modelling. A number of variants of the MsFV
method have been proposed since its introduction
in~\citep{jenny2003multi}. In this manuscript, we employ the MsFV
method as described in~\citep{lunati2006multiscale, lunati2008multiscale}.

\subsection{Multiscale finite volume method}

\begin{figure}
  \centering
  \begin{tikzpicture}[scale=.5]
    \draw[step=1,gray,very thin] (0,0) grid +(15,15);
    \draw[step=5,black,thick] (0,0) grid +(15,15);
    \draw[step=5,blue,thin,dashed,shift={(2.5,2.5)}] (-2.5,-2.5) grid +(15,15);
    \fill[green!50!black,opacity=0.3] (0,10) rectangle +(5,5);
    \fill[orange!60!black,opacity=0.3] (7,2) rectangle +(6,6);
    \foreach \x in {2.5,7.5,12.5} {
      \foreach \y in {2.5,7.5,12.5} {
        \node [text=red] at (\x,\y) {$\bullet$};
      }
    }
    \node at (2.5,12.5) [above right] {$\Omega^C_i$};
    \node at (10,5) [above right] {$\Omega^D_j$};
  \end{tikzpicture}
  \caption{A square domain with a fine discretization of size
    $15\times 15$ (grey thin lines). A coarse \emph{primal grid} of
    size $3\times 3$ is defined on top of the fine grid (black bold
    lines). A \emph{primal cell} $\Omega^C_i$ is highlighted in
    green. The centers of each primal cell are marked with red
    dots. From these centers, the \emph{dual grid} is defined (blue
    dashed lines). A \emph{dual cell} $\Omega^D_j$ is highlighted in
    light red.}
  \label{fig:grids}
\end{figure}
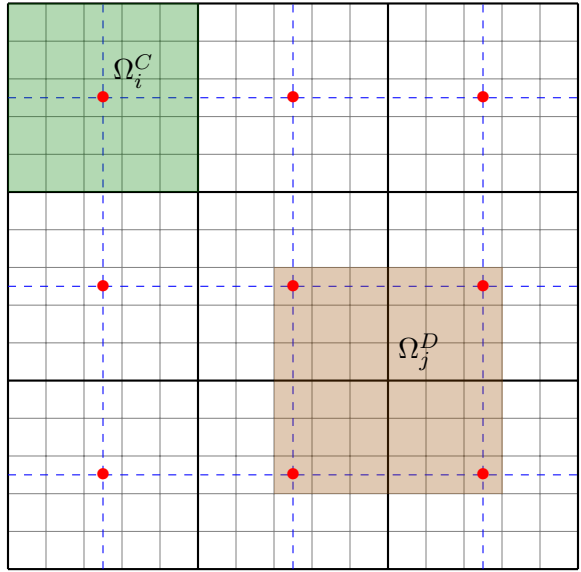

\begin{figure}
  \centering
  \begin{tikzpicture}[scale=.5]
    \draw[step=1,gray,very thin] (0,0) grid +(15,15);
    \draw[step=5,black,semithick] (0,0) grid +(15,15);
    \draw[step=5,blue,thin,dashed,shift={(2.5,2.5)}] (-2.5,-2.5) grid +(15,15);
    \fill[blue!50,opacity=0.3] (2,2) rectangle +(11,11);
    \foreach \y [count=\yi] in {2.5,7.5,12.5} {
      \foreach [count=\xi, evaluate=\xi as \c using int((\yi-1)*3+\xi)] \x in {2.5,7.5,12.5} {
        \node [text=red] at (\x,\y) {$\bullet$};
        \node [text=black] at (\x,\y) [above left] {$\c$};
      }
    }
    \fill[green!50!black,opacity=0.5] (7,7) rectangle +(1,1);
  \end{tikzpicture}
  \caption{Shown in blue is the support region of basis function
    $\phi^5$ corresponding to coarse node $5$ (green cell). Basis
    function $\phi^5$ is obtained by solving the local
    problems in Eq.~\eqref{eq:local_prob} for $i=5$, then
    $\phi^5 = \sum^{N_D}_{j=1}{\phi^5_j}$. In this example,
    only $\phi^5$ is an \emph{interior} basis function (see Section~\ref{methodology}).}
  \label{fig:supportregion}
\end{figure}
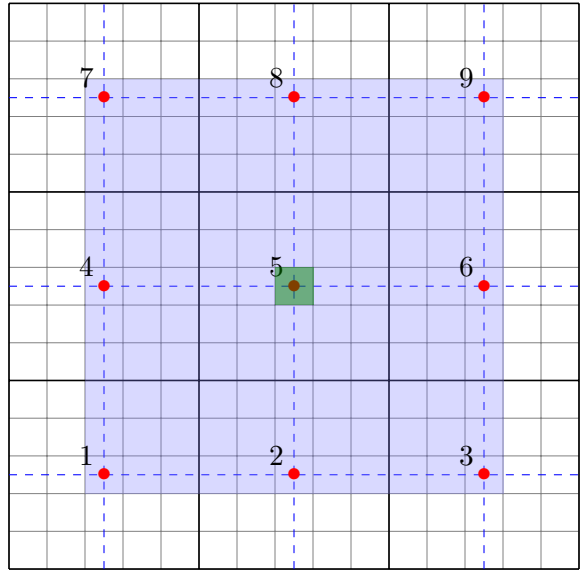

\begin{figure}
  \centering
  \begin{subfigure}{.49\textwidth}
    \centering
    \includegraphics[width=\textwidth]{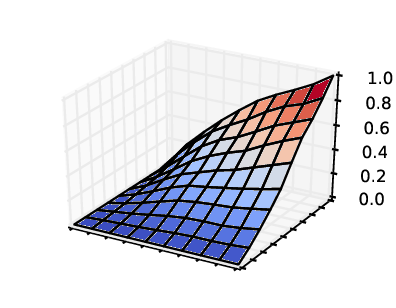}
    \caption{A partial basis function.}
    \label{fig:partial_basis_surf}
  \end{subfigure}
  \begin{subfigure}{.49\textwidth}
    \centering
    \includegraphics[width=\textwidth]{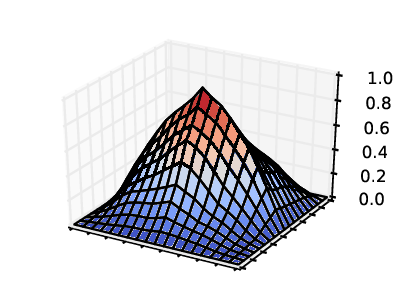}
    \caption{A complete basis function.}
    \label{fig:basis_surf}
  \end{subfigure}
  \caption{Illustration of a local solution (partial basis function)
    and a basis function within its support region.}
  \label{fig:basis_function}
\end{figure}

We consider an elliptic equation describing the pressure field
\begin{equation}
  \label{eq:pressure}
  -\nabla\cdot(\mathbf{K}\nabla p) = q
\end{equation}
where $p$ denotes the fluid pressure, $q$ denotes fluid sources, and
$\mathbf{K}$ denotes the permeability tensor. Discretizing
Eq.~\eqref{eq:pressure} by the finite volume method results in a
system of equations of the form $\mathbf{A}\mathbf{p}=\mathbf{q}$,
which for some applications (such as reservoir simulation) tends to be
extremely large. The MsFV method tackles this problem by constructing
and solving a much coarser system of equations
$\mathbf A_C \mathbf p_C = \mathbf r_C$,
the solution of which is then used to obtain an approximation of
$\mathbf{p}$ by interpolation. For this, it relies on a set of basis
functions which are obtained by solving local problems. In this sense,
the method slightly resembles the finite element method, except that
the basis functions employed are not analytic functions but
numerically computed functions.

The method begins with the definition of a pair of overlapping coarse
grids, namely the \emph{primal grid} and the \emph{dual grid}, as
shown in Figure~\ref{fig:grids}. In principle, the primal grid can
be any coarse partition defined over the fine grid. Next, we define the
\emph{coarse nodes} as the fine cells at the centers of each primal
cell. Lastly, the dual grid is defined by the lines connecting these
coarse nodes. We denote the primal cells with
$\Omega^C_i, \; i\in\{1,\cdots,N_C\}$, and the dual cells with
$\Omega^D_j, \; j\in\{1,\cdots,N_D\}$.

A set of (partial) basis functions are obtained by solving the local
problems
\begin{align}\label{eq:local_prob}
  \begin{split}
  \nabla\cdot (\mathbf{K}\cdot\nabla\phi^i_j) &= 0 \quad \text{in} \; \Omega^D_j \\
  \nabla\cdot{}_{\lVert}(\mathbf{K}\cdot\nabla\phi^i_j){}_{\rVert}&= 0 \quad \text{on} \; \partial\Omega^D_j \\
\phi^i_j(\mathbf x_k) &= \delta_{ik} \quad k \in \{1, \cdots, N_C\},
\end{split}
\end{align}
where $\phi^i_j$ denotes the (partial) basis function on dual cell
$\Omega^D_j$ (see Figure~\ref{fig:partial_basis_surf}) associated with
coarse node $i$, $\mathbf x_k$ denotes the coordinate of coarse node
$k$, and ${}_{\lVert}\cdot{}_{\rVert}$ denotes the tangential
component over the dual cell boundary $\partial\Omega^D_j$. In the 2D
case, this means solving 1D problems over $\partial\Omega^D_j$, the
solutions of which become the boundary conditions for the 2D problem on
$\Omega^D_j$.

Another component of the MsFV formulation employed are the correction
functions, obtained by solving the local problems
\begin{align}\label{eq:local_corr}
  \begin{split}
  \nabla\cdot (\mathbf{K}\cdot\nabla\hat{\phi}_j) &= q \quad \text{in} \; \Omega^D_j \\
  \nabla\cdot{}_{\lVert}(\mathbf{K}\cdot\nabla\hat{\phi}_j){}_{\rVert}&= q \quad \text{on} \; \partial\Omega^D_j \\
  \hat{\phi}_j(\mathbf x_k) &= 0 \quad k \in \{1, \cdots, N_C\},
  \end{split}
\end{align}
where $\hat{\phi}_j$
denotes the correction function on dual cell $\Omega^D_j$.

Once the basis and correction functions are obtained, we approximate
the fine scale pressure $p$ as an interpolation of coarse scale
pressure values $\mathbf p_C$ plus a correction term
\begin{equation}\label{eq:interpol}
  p \approx
  \sum^{N_D}_{j=1}{(\sum^{N_C}_{i=1}{\phi^i_j p^i_C} + \hat{\phi}_j)}
\end{equation}
Substituting Eq.~\eqref{eq:interpol} in Eq.~\eqref{eq:pressure}, and applying
finite volume discretization over the primal grid, we get the coarse
system of equations $\mathbf A_C \mathbf p_C = \mathbf r_C$.

In the current work, it is more convenient to express the
basis and correction functions in a global point of view (see
Figure~\ref{fig:basis_surf}) by writing
$\phi^i = \sum^{N_D}_{j=1}{\phi^i_j}$ and
$\hat{\phi} = \sum_{j=1}^{N_D} \hat{\phi}_j$
(for the basis functions, we actually only need to sum over supporting
dual cells, i.e. dual cells that are associated with the corresponding
node). Then, Eq.~\eqref{eq:interpol} has a simpler interpolation
expression of the form:
\begin{equation}\label{eq:interpol2} 
  p \approx \sum^{N_C}_{i=1}{\phi^i p^i_C} + \hat{\phi}
\end{equation}

In the case that the pressure solution will be utilised to drive a
transport problem at the fine scale, a flux reconstruction step
consisting of solving additional local Neumann problems is necessary
\citep{lunati2006multiscale}.

\subsection{Feedforward neural networks for surrogate modelling} 

\tikzset{%
  every neuron/.style={
    circle,
    draw,
    minimum size=1cm
  },
  neuron missing/.style={
    draw=none, 
    scale=4,
    text height=0.333cm,
    execute at begin node=\color{black}$\vdots$
  },
}

\begin{figure}[t]
  \centering
\begin{tikzpicture}[x=1.5cm, y=1.5cm, >=stealth]

\foreach \m/\l [count=\y] in {1,2,missing,3}
  \node [every neuron/.try, neuron \m/.try] (input-\m) at (0,2.5-\y) {};

\foreach \m [count=\y] in {1,2,3,missing,4}
  \node [every neuron/.try, neuron \m/.try ] (hidden-\m) at (2,3.0-\y*1.0) {};

\foreach \m [count=\y] in {1,2,missing,3}
  \node [every neuron/.try, neuron \m/.try ] (output-\m) at (4,2.5-\y) {};

\foreach \l [count=\i] in {1,2,d_{in}}
  \draw [<-] (input-\i) -- ++(-1,0)
    node [above, midway] {$x_{\l}$};

\foreach \l [count=\i] in {1,n}
  \node [above] at (hidden-\i.north) {};

\foreach \l [count=\i] in {1,2,d_{out}}
  \draw [->] (output-\i) -- ++(1,0)
    node [above, midway] {$y_{\l}$};

\foreach \i in {1,...,3}
  \foreach \j in {1,...,4}
    \draw [->] (input-\i) -- (hidden-\j);

\foreach \i in {1,...,4}
  \foreach \j in {1,...,3}
    \draw [->] (hidden-\i) -- (output-\j);

\foreach \l [count=\x from 0] in {Input, Hidden, Output}
  \node [align=center, above] at (\x*2,2.35) {\l \\ Layer};

\end{tikzpicture}
\caption{Representation of a 1-hidden layer neural network as a
  graph. The first column of nodes (from left to right) is the input
  layer, taking inputs $\mathbf x=(x_1,\cdots,x_{d_{in}})$ of
  dimension $d_{in}$, and the last column is the output layer with
  output $\mathbf y=(y_1,\cdots,y_{d_{out}})$ of dimension
  $d_{out}$. The intermediate column is the hidden layer. Each line
  connecting two nodes represents a multiplication by a scalar weight.
  }
  \label{fig:nn}
\end{figure}
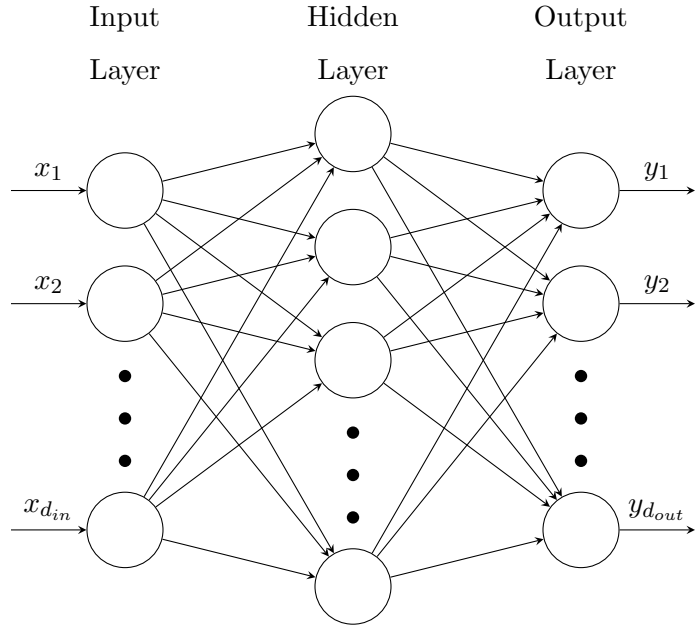

A \emph{feedforward neural network} $f$ is
a function consisting of a compositional chain of simpler functions,
i.e.
$f(\mathbf x)=f^{(n)}(f^{(n-1)}(\cdots(f^{(1)}(\mathbf x))\cdots))$.
Here, $f$ is said to have $n$ \emph{layers}, where $f^{(1)}$ is the
first layer, $f^{(2)}$ the second layer, etc.  The first layer is also
called the \emph{input layer}, the last layer the \emph{output layer},
and the layers in between are called \emph{hidden layers}.
The size or \emph{number of units} of a layer
$f^{(i)}:\mathbb R^{d_{in}^i} \to \mathbb R^{d_{out}^i}$
refers to its output dimension, i.e. $d_{out}^i$.
Neural networks are typically represented as a graph
as shown in Figure~\ref{fig:nn}.

The input layer encompasses any pre-processing of the data, while the
output layer may take many forms depending on the learning task. For
example, in classification tasks, the \emph{softmax} function is
normally employed where the output layer size (number of units) equals
the number of classes, and the softmax output represents the
probability of the input sample belonging to each class. This is a way
of embedding prior information of the learning task into the
model. For regression tasks, the identity function is normally
employed, but other functions can also be chosen to embed any prior
information of the output space.

A conventional hidden layer is an affine transformation followed by an
element-wise nonlinear function or \emph{activation}, i.e.
$f^{(i)}(\mathbf x) = \sigma_i(\mathbf W_i\mathbf x+\mathbf b_i)$.
Some popular choices for $\sigma_i$ are the hyperbolic tangent, the sigmoid
function, and more recently, the rectified linear unit~\citep{jarrett2009best}.

\subsubsection{Neural network optimization}
Once the network \emph{architecture} is defined (number of layers $n$,
layer sizes $d_{out}^i$, activation functions $\sigma_i$, etc.),
the optimization problem is to find
$\boldsymbol\theta=[\mathbf W_1;\mathbf b_1,\cdots,\mathbf W_n;\mathbf
b_n]$ such that $f$ best describes the observations. Let
$\{(\mathbf x_1,\mathbf y_1),\cdots,(\mathbf x_{N}, \mathbf
y_{N})\}$ be the set of observations used to train the model
(the \emph{training set}), then the minimization problem is formulated as:
\begin{equation}
  \label{eq:argmin} \operatorname*{arg\,min}_{\boldsymbol\theta}{
\frac{1}{N}\sum^{N}_{i=1}||\mathbf y_i-f(\mathbf x_i;\boldsymbol\theta)||^2_2}
\end{equation}
where
$J(\boldsymbol\theta)\coloneqq
\frac{1}{N}\sum^{N}_{i=1}||\mathbf y_i-f(\mathbf x_i;\boldsymbol\theta)||^2_2$
is the \emph{cost function}. This problem can
be solved using gradient-based algorithms such as gradient descent:
\begin{equation}
\label{eq:gradient_descent}
\boldsymbol\theta^{k+1} = \boldsymbol\theta^k - \epsilon \nabla_{\boldsymbol\theta}J(\boldsymbol\theta)
\end{equation}
where $\epsilon$ is the \emph{learning rate}. The derivative
$\nabla_{\boldsymbol\theta}J$ can be obtained with numerical
differentiation schemes. In neural networks, this is typically the
\emph{backpropagation} algorithm~\citep{rumelhart1988learning}.

More recent gradient-based algorithms improve over gradient descent by
offering adaptive learning rates such as
AdaGrad~\citep{duchi2011adaptive}, RMSProp~\citep{tieleman2012lecture}
and Adam~\citep{kingma2014adam}. The basic idea in these methods is to
use a separate learning rate for each scalar parameter, and adapt
these rates throughout the training process based on the historical
values of the partial derivatives with respect to each parameter. The
initial global learning rate $\epsilon_{0}$ is a tunable
hyperparameter.

A recent important development regarding neural network optimization is
\emph{batch normalization}~\citep{ioffe2015batch}, which has shown to
significantly speed up the optimization process. The method consists
of adaptive reparametrization of inputs to each activation
function. This is to address the simplifying assumptions made during
the optimization where the gradients update each parameter assuming
that the other layers do not change. In essence, the values after the
affine transformation in a layer are normalized by the mean and
standard deviation before being fed into the layer activation
function.

\subsubsection{Regularization} \label{sec:regularization}
The optimization of $\boldsymbol\theta$ from
Eq.~\eqref{eq:argmin} alone yields a model that is prone to
overfitting, i.e. it does not necessarily perform well for samples not
seen in the training set.
Hence, validation assessment is necessary where a separate set of
samples that are not used in optimizing Eq.~\eqref{eq:argmin},
called the \emph{validation set}, is employed to assess the accuracy of
the model. 
One simple regularization technique is \emph{early stopping}, where
the model is assessed after each update (or number of updates) of
$\boldsymbol\theta$ in Eq.~\eqref{eq:gradient_descent}; when the
cost function on the validation set begins to increase, the
optimization is early stopped. This is the stopping criteria
in neural network optimization, which differs from conventional
optimization where the criteria is generally based on the
gradient norm.
Another set of regularization techniques are \emph{parameter norm penalties}, where an
additional term is added to the cost function:
\begin{equation}
J_{reg}(\boldsymbol\theta)\coloneqq
\frac{1}{N}\sum^{N}_{i=1}||\mathbf y_i-f(\mathbf x_i;\boldsymbol\theta)||^2_2
+\alpha\Omega(\boldsymbol\theta)
\end{equation}
For example, for $L^2$ parameter norm, 
$\Omega(\boldsymbol\theta)= \frac{1}{2}||\boldsymbol\theta||^2_2$.
The additional parameter $\alpha$ is a regularization \emph{hyperparameter}
that is chosen using the validation set.

More recently, \emph{Dropout}~\citep{srivastava2014dropout} has shown
to be a very effective regularization technique that approximates
model averaging in neural networks. The technique consists of randomly
dropping out units of the network during the optimization iteration,
by which the optimizer ``sees'' a number of different ``models'' in
the process. Since traditional model averaging in neural networks is
usually intractable, this approach serves as a proxy for averaging
an exponential number of models. In practice, a \emph{dropout rate} is
chosen which indicates the probability of a unit being dropped
out. This hyperparameter is tuned using a validation set. 
The authors in \citep{srivastava2014dropout} suggested
using a \emph{max-norm} constraint along with dropout, which 
consists of constraining the norm of some of the weights by a fixed
constant $c$, tuned with a validation set. This allows for more
aggressive optimization search without the possibility of weights
blowing up.

\subsubsection{Architecture design and hyperparameter tuning} \label{sec:arch_design}
The design of the network architecture, i.e. defining the number of
layers, the number of units for each layer, activation functions,
regularizers, etc. is not a straightforward task. In principle, one
can consider these parameters as additional hyperparameters and tune
them using a validation set. However, hyperparameter optimization is
an expensive task given that the objective function (performance on
the validation set) is non-linear and non-differentiable with respect
to the hyperparameters. 
In practice, heuristics and expertise are heavily employed
in the design process to reduce the number of hyperparameters. 
Nevertheless, general guidelines do exist for the design of neural
network models. The following is a compilation of guidelines extracted
from~\citep{Goodfellow-et-al-2016,bengio2012practical}:
\begin{itemize}
\item[--] Begin with a few number of layers and units, and well-tested
optimizers and regularizers.
\item[--] Start with as few hyperparameters as possible to enable quick
  manual search to obtain some insight of the learning task.
\item[--] Overfit, then regularize: increase the number of
  layers/units to overfit the training set, then apply
  regularization techniques to improve generalization. That is, we
  first want the network to be complex enough to approximate
  the target function, and then regularize it to perform well
  outside the training set.
\item[--] Regarding the choice of activation functions, the current
  default recommendation is to use rectified linear units (ReLUs). These
  have beneficial properties for the optimization such as non-vanishing
  gradients.
\item[--] Whether to use few large layers, or many small layers is an
  open debate. It is generally believed that many small layers
  generalize better, although the ultimate decision will largely depend on
  implementation and trial and error.
\item[--] Early stopping should almost always be employed.
\item[--] The learning rate $\epsilon_0$ is very influential in the
  model performance and should be fine-tuned.
\item[--] If there is an architecture that performs well for a similar
  task, use it as the base architecture.
\end{itemize}

Once the general architecture is selected, key hyperparameters should
be optimized. Traditional techniques such as grid search
and random search are normally prohibitive. In the current manuscript
we employed the \emph{Tree-Parzen
  Estimator}~\citep{bergstra2011algorithms}, a 
sequential model-based hyperparameter optimization approach where a model is sequentially constructed to
approximate the performance of hyperparameters based on historical
measurements, and then subsequently choose new hyperparameters to test
based on this model. Other hyperparameter optimization techniques include Bayesian
optimization for neural networks~\citep{snoek2012practical} and
Hyperband~\citep{li2016hyperband}.
\section{Methodology} \label{methodology}
\begin{figure}
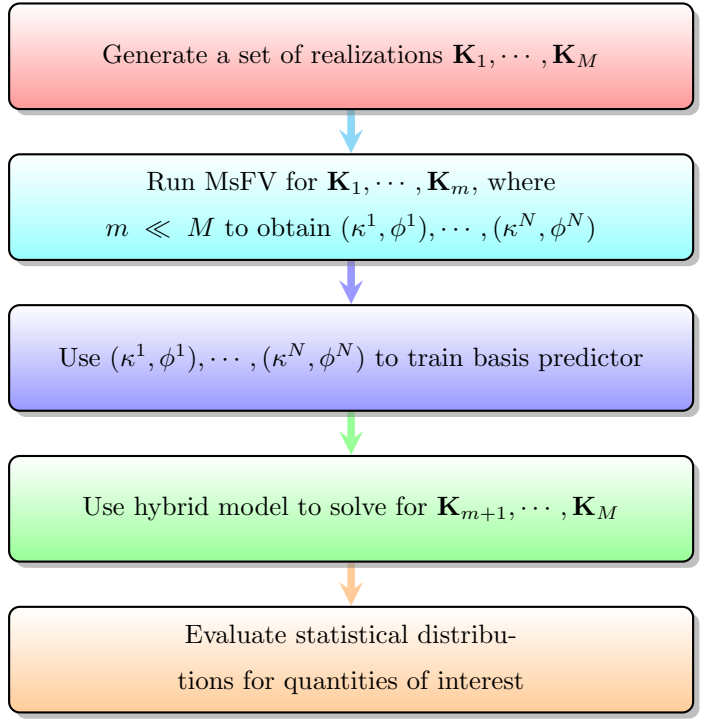

  \centering
\smartdiagramset{border color=black, module y sep=2.0, back arrow disabled=true,
module minimum height= 1.4cm, text width = 8cm, module minimum width=9cm}
  \smartdiagram[flow diagram] { 
    {Generate a set of realizations $\mathbf K_1, \cdots, \mathbf K_M$},
    {Run MsFV for {$\mathbf K_1, \cdots, \mathbf K_m$}, where $m \ll M$ to
    obtain {$(\kappa^1, \phi^1), \cdots, (\kappa^N, \phi^N)$}},
    {Use {$(\kappa^1, \phi^1), \cdots, (\kappa^N, \phi^N)$} to train basis predictor},
    {Use hybrid model to solve for {$\mathbf K_{m+1}, \cdots, \mathbf K_M$}},
    {Evaluate statistical distributions for quantities of interest}
  }
  \caption{Workflow of the proposed method.}
  \label{fig:workflow}
\end{figure}

We first introduce some terminology to simplify the presentation. A
basis function is \emph{interior} if its support region does
\emph{not} touch the domain boundary (see
Figure~\ref{fig:supportregion}). For practical purposes, we limit the
learning process to interior basis functions, i.e. we build a
predictor to generate interior basis functions while computing the
remaining basis functions as usual from the local problems defined by
Eq.~\eqref{eq:local_prob}. In practice, this should not be a major
concern given that the number of interior basis functions is normally
much larger than the number of non-interior basis functions (basis
functions on the \emph{edges} and \emph{vertices} of the domain).
In any case, it is pretty straightforward to train additional
predictors for the remaining types of basis functions.

We define a \emph{permeability patch} $\kappa^i$ as the cropped region
of the permeability field $\mathbf K$ that corresponds to the support
region of a basis function $\phi^i$. In the learning framework, the
permeability patches $\kappa^1, \kappa^2, \cdots, \kappa^N$ are our
inputs, and the corresponding basis functions
$\phi^1, \phi^2, \cdots, \phi^N$ are our outputs. In practice, it is
more convenient to work with the log-permeabilities,
i.e. $\log\kappa^1,\cdots\log\kappa^N$.

For clarification, suppose we have an $81\times 81$
Cartesian grid, over which we defined a $9\times 9$ primal grid. Then
there are $7\times7=49$ interior basis functions, each with array size
(number of fine cells in the support region) of $19\times 19$. The
array size of each input permeability patch is $19\times 19 \times d$,
where $d=1$ for isotropic fields, and $d=2$ in the anisotropic case.

The method we propose aims to speedup uncertainty quantification
studies where multiscale methods are employed in the propagation
task. In particular, we focus on the use of the MsFV method to solve
Eq.~\eqref{eq:pressure} for large number of realizations of the
permeability field $\mathbf{K}$. Our method however could be applied
to any multiscale method where the explicit generation of the basis
functions is performed. Consider an uncertainty propagation task of
solving Eq.~\eqref{eq:pressure} for~$\mathbf K_1, \cdots, \mathbf
K_M$. Let $m\ll M$ be the number of full MsFV runs that we can
afford. For each $\mathbf K_i$ where $i=1,\cdots,m$, the MsFV simulation
delivers a set of basis functions with their corresponding permeability
patches. The union of these sets provides the learning
dataset $\{(\kappa^1, \phi^1), \cdots, (\kappa^N, \phi^N)\}$. This
dataset is used to train a predictor model that maps from permeability
patches $\kappa$ to basis functions $\phi$ that has an evaluation cost
that is much cheaper than solving the local
problems. 
The model we employed here is a neural network.
Once the model is trained, it is used to predict the basis
functions in the subsequent runs for~$\mathbf K_{m+1}, \mathbf
K_{m+2}, \cdots, \mathbf K_M$. The end result is a hybrid approach
where the MsFV formulation is modified to obtain the basis functions
through a data-driven model instead of being computed from the local
problems. 
Figure~\ref{fig:workflow} summarises the workflow of the proposed method.

To ensure the partition of unity property of the basis functions
(which is not necessarily fulfilled in the learning model), we
perform a post-processing step on the generated basis functions
as follows:
\begin{equation*}
\phi^i_{new}(x) = \frac{\phi^i(x)}{\sum_k \phi^k(x)}
\end{equation*}

We note that the presented method benefits from the use
of structured grids with coarse cells of same size. In the case of
structured grids with cells of different sizes, the patches could be
scaled to a unique input size. This is a standard preprocessing step
in computer vision to handle images of different sizes. We also note
that handling unstructured grids is not a straightforward task and
is beyond the scope of the current manuscript.

\subsection{Implementation and computational aspects}
The computational gain of the proposed method is achieved by replacing
the solution of local problems by a constant number of matrix-vector
multiplications followed by element-wise function evaluations. For a
network of input and output sizes $n$ and hidden layers of size $m$,
this means an evaluation complexity of $\mathcal{O}(mn)$.
The constants associated with the evaluation cost will depend on the
number of layers of the network.

Efficient algorithms for solving systems of $n$ linear equations exist
where complexities of $\mathcal{O}(n\log n)$ (FFT) or even
$\mathcal{O}(n)$ (multigrid methods) are achieved, however the
constants associated with these algorithms are large, usually
requiring a large value of $n$ to be economical. This is in contrast
to the MsFV approach where small local problems are
preferred. Likewise, there exist efficient algorithms to perform
matrix-vector multiplications that are only justified in practice
when the matrices and vectors are very large.

Regarding the training of the neural network, this is performed in an
\emph{offline phase} as with other surrogate modelling techniques, and
the justification of the cost will depend on the particular
uncertainty quantification task at hand. The larger the uncertainty
quantification task, the larger the time budget that can be assigned
to the surrogate modelling process.  As a practical note, it is worth
mentioning that due to the surge of interest in neural networks
and AI in general, efficient implementations have been intensively
developed in recent years, both from the software and hardware
sectors. Indeed, dedicated hardware devices are currently being
released for various neural network implementations.

 \subsection{Other machine learning techniques}
The proposed algorithm is not limited to neural networks and other
traditional machine learning techniques are indeed applicable such as
Gaussian processes and support vector machines to model the basis
function predictor. A number of reasons led us to choose the neural
network model. First, neural networks are universal
approximators~\citep{hornik1989multilayer}, meaning that they can fit
any measurable function with arbitrary accuracy. This practically covers
any function encountered in engineering
applications. Secondly, neural networks scale very well with the size
of the dataset, in contrast to Gaussian processes and support vector
machines. This is desirable where the trends of the cost of numerical
simulations and data storage are ever decreasing. Lastly, research in
neural networks is characterized by a remarkably large and evolving
body of literature from which we can benefit in the near
future. Advances in the field is specially strong in problems
related to computer vision, which shares a lot of features
with our work.
\section{Numerical experiments}
We consider the task of solving Eq.~\eqref{eq:pressure} over a unit
square $[0,1]^2\subset \mathbb{R}^2$. 
The domain is discretized into
$81\times81$ fine grid, with a primal coarse grid of
$9\times 9$. 
For estimating the statistical distributions, we utilize
$M=1000$ realizations of isotropic log-permeability fields generated
assuming a zero mean gaussian random field with an exponential
covariance of the form
\begin{equation*}
\operatorname{Cov}(\mathbf x_1, \mathbf x_2) = \sigma^2\exp\left(-\frac{\lVert
    \mathbf x_1-\mathbf x_2 \rVert}{L}\right)
\end{equation*}
where $\lVert\cdot\rVert$ denotes the Euclidean norm.
We choose $\sigma=1.0$, and we investigate three values for the
correlation length: $L=0.1$, $0.2$, and $0.4$.

\subsection{Learning process} \label{sec:learning_process}
\begin{table}
  \centering
  \caption{$R^2$-scores on different permeability types}
  \begin{tabular}{lr}
    \hline
    Correlation length & $R^2$-score \\
    \hline
    $L=0.1$ & $0.927$\\
    $L=0.2$ & $0.953$\\
    $L=0.4$ & $0.964$\\
    \hline
  \end{tabular}
  \label{table:r2scores}
\end{table}

\begin{figure}
  \centering
  \begin{subfigure}{\textwidth} \centering
  \includegraphics[width=.7\textwidth]{{{0.1_input}}}
  \caption{Input $\log\kappa$}
  \end{subfigure}
  \begin{subfigure}{\textwidth} \centering
  \includegraphics[width=.7\textwidth]{{{0.1_target}}}
  \caption{Target $\phi$}
  \end{subfigure}
  \begin{subfigure}{\textwidth} \centering
  \includegraphics[width=.7\textwidth]{{{0.1_predicted}}}
  \caption{Predicted $\hat\phi$}
  \end{subfigure}
  \begin{subfigure}{\textwidth} \centering
  \includegraphics[width=.7\textwidth]{{{0.1_difference}}}
  \caption{Difference $\phi - \hat\phi$}
  \end{subfigure}
  \caption{Performance of basis function predictor, case $L=0.1$}
  \label{fig:predictor_01}
\end{figure}

\begin{figure}
  \centering
  \begin{subfigure}{\textwidth} \centering
  \includegraphics[width=.7\textwidth]{{{0.2_input}}}
  \caption{Input $\log \kappa$}
  \end{subfigure}
  \begin{subfigure}{\textwidth} \centering
  \includegraphics[width=.7\textwidth]{{{0.2_target}}}
  \caption{Target $\phi$}
  \end{subfigure}
  \begin{subfigure}{\textwidth} \centering
  \includegraphics[width=.7\textwidth]{{{0.2_predicted}}}
  \caption{Predicted $\hat\phi$}
  \end{subfigure}
  \begin{subfigure}{\textwidth} \centering
  \includegraphics[width=.7\textwidth]{{{0.2_difference}}}
  \caption{Difference $\phi - \hat\phi$}
  \end{subfigure}
  \caption{Performance of basis function predictor, case $L=0.2$}
  \label{fig:predictor_02}
\end{figure}

\begin{figure}
  \centering
  \begin{subfigure}{\textwidth} \centering
  \includegraphics[width=.7\textwidth]{{{0.4_input}}}
  \caption{Input $\log \kappa$}
  \end{subfigure}
  \begin{subfigure}{\textwidth} \centering
  \includegraphics[width=.7\textwidth]{{{0.4_target}}}
  \caption{Target $\phi$}
  \end{subfigure}
  \begin{subfigure}{\textwidth} \centering
  \includegraphics[width=.7\textwidth]{{{0.4_predicted}}}
  \caption{Predicted $\hat\phi$}
  \end{subfigure}
  \begin{subfigure}{\textwidth} \centering
  \includegraphics[width=.7\textwidth]{{{0.4_difference}}}
  \caption{Difference $\phi - \hat\phi$}
  \end{subfigure}
  \caption{Performance of basis function predictor, case $L=0.4$}
  \label{fig:predictor_04}
\end{figure}

We assume a budget of $m=20$ full MsFV runs, obtaining a dataset of
$980$ samples (since each run yields 49 samples). The array sizes of
the inputs (permeability patches) and outputs (basis functions) are
$19\times 19$.
We set aside $20\%$ of the dataset for validation (this should be done
at the level of the realizations, i.e. samples generated from $4$ full
MsFV runs).

The architecture employed is a fully connected network with 1-hidden
layer of size 1024 and ReLU activation function. Naturally, the input
and output layers are of size $19\times 19$ = 361, matching the size
of the permeability patch and basis function. Additionally, we employ
the hard sigmoid function as the activation of the output layer. This
is to embed the prior knowledge that basis functions take values
between $0$ and $1$. The hard sigmoid is the function
${x\in\mathbb{R} \mapsto \max(0,\min(1,0.2x+0.5))}$. Although this
choice of function is usually problematic for gradient-based
optimizations, it gave good results when coupled with dropout and
batch normalization. 

To train the network, the gradient-based optimizer
Adam~\citep{kingma2014adam} seemed more robust during our trials. The
initial learning rate was set to $\epsilon_0=10^{-3}$. For regularization, a
dropout rate of $5\%$ after the hidden layer, and a max-norm constraint
of $4$ have proven useful. Additionally, early
stopping is employed. All these hyperparameter values were chosen
based on default recommendations along with some manual explorations.

A convenient metric employed to report the performance of a trained
model is the coefficient of determination (or $R^2$-score):
\begin{equation*}
R^2(f) = 1 -
\frac{\sum{{\lVert\phi^i-\hat{\phi}^i\rVert}^2}}{\sum{{\lVert\phi^i-\bar{\phi}\rVert}^2}}
\end{equation*}
where $\lVert\cdot\rVert$ denotes the $L^2$ norm, $f$ is the trained
model, $\hat\phi^i=f(\log\kappa^i),\; i=1,2,\cdots,N_{val}$ are the
\emph{predicted} basis functions,
$\phi^1,\phi^2,\cdots,\phi^{N_{val}}$ are the \emph{true} basis
functions, and $\bar\phi=\frac{1}{N_{val}}\sum\phi^i$ is the sample
mean of the true basis functions. A score of $1.0$ corresponds to
perfect prediction, while a score below $0$ means that the predictor
performance is worse than a model that always predicts the sample
mean.
Table~\ref{table:r2scores} shows the validation scores obtained
on the three permeability types considered, i.e. correlation lengths
$L=0.1$, $0.2$ and $0.4$. Figures~\ref{fig:predictor_01},
\ref{fig:predictor_02} and~\ref{fig:predictor_04} show some of the
predicted basis functions for cases $L=0.1$, $0.2$ and $0.4$,
respectively. We see that the prediction is more challenging for the
case of shortest correlation length. This is likely due to the
permeability field being more heterogeneous for shorter correlation
lengths.

\subsubsection{Predictor uncertainty} Error estimations of the
predicted basis functions might be of interest to fully quantify the
uncertainties in the results.  Such estimations are readily
available in machine learning models such as Gaussian processes. For
neural networks, a number of
methods such as Bootstrap aggregating (bagging) and others as discussed
in~\citep{tibshirani1996comparison} could be employed.
Another possibility is to employ
the dropout technique as a Bayesian
approximation method~\citep{gal2016dropout}. In our work, we consider the
uncertainties in the predicted basis functions to be of second order
and the presented numerical results support this assumption.

\subsection{Hybrid model}

Once the neural network model is trained, we can use it to compute the basis
functions in the MsFV formulation. To assess the effectiveness of this
hybrid approach (MsFV-NN), we consider two test cases:

\setlist[description]{font=\normalfont\itshape\space}

\begin{description}
\item[Quarter-five spot problem:]
  In this problem, injection and production points are located at
  $(0,0)$ and $(1,1)$ of the unit square, respectively.  No-flow
  boundary conditions are imposed. We assume unit injection/production
  rates, i.e. $q(0,0)=1$ and $q(1,1)=-1$.

\item[Uniform flow problem:]
  Here, uniformly distributed inflow and outflow conditions are imposed
  on the left and right sides of the unit square, respectively. No-flow
  boundary conditions are imposed on the remaining top and bottom
  sides. A total inflow/outflow rate of $1$ is assumed. For the unit
  square, this means $v\cdot\hat{n}=-1$ and $v\cdot\hat{n}=1$ on the
  left and right sides, respectively, where $\hat{n}$ denotes the
  outward-pointing unit normal to the boundary.
\end{description}

In both cases, a pressure value of $0$ is imposed at the center of the
square to close the problem. The pressure Eq.~\eqref{eq:pressure} is
solved using three methods: a standard cell-centered finite volume
method at the fine-scale level which is taken as the reference
``true'' solution, the standard MsFV method, and the proposed hybrid
method (MsFV-NN). Additionally, we also compute and compare the total
velocities, which can be derived from the corresponding pressure
solutions. In the reference solution, the total velocity can be
derived using Darcy's law ($v=-\lambda\mathbf K\nabla p$ where
$\lambda$ is the total mobility, here assumed as $\lambda=1$), whereas
in the MsFV and MsFV-NN methods, the total velocities are derived via
a flux reconstruction step, as mentioned before. We take a further
step and use the total velocity to solve a tracer flow problem. In
this case, we solve the following advection equation:
\begin{equation}\label{eq:concentration}
\varphi\frac{\partial c}{\partial t} + \nabla\cdot(cv) =
\frac{q_w}{\rho_w}
\end{equation}
where $c$ denotes the concentration of the injected fluid (in this case
water), $\varphi$ denotes the domain porosity, $q_w$ denotes
sources/sinks of the injected fluid, and $\rho_w$ denotes the density
of the injected fluid. In all cases, we assume water with
dimensionless density of $\rho_w=1$ that is injected into a reservoir
with constant porosity $\varphi=0.2$ initially containing only oil,
i.e.  $c(\mathbf x,t=0) = 0$, which we assume to have the same
viscosity as the injected fluid. Under these conditions the total
velocity $v$ does not change in time.  The simulation time for both
test cases is from $t=0$ until $t=0.4$. In reservoir engineering, it
is more convenient to work with pore volume injected (PVI) as the time
unit, which expresses the ratio of the total volume of fluid injected
until time $t$ and the reservoir pore volume (for constant injection,
$t_{PVI}=q_{in}t/V_{\varphi}$ where $V_{\varphi}$ is the pore volume).

\begin{figure}
  \centering
  \begin{subfigure}{\linewidth}
  \centering
    \includegraphics[height=.45\textheight]{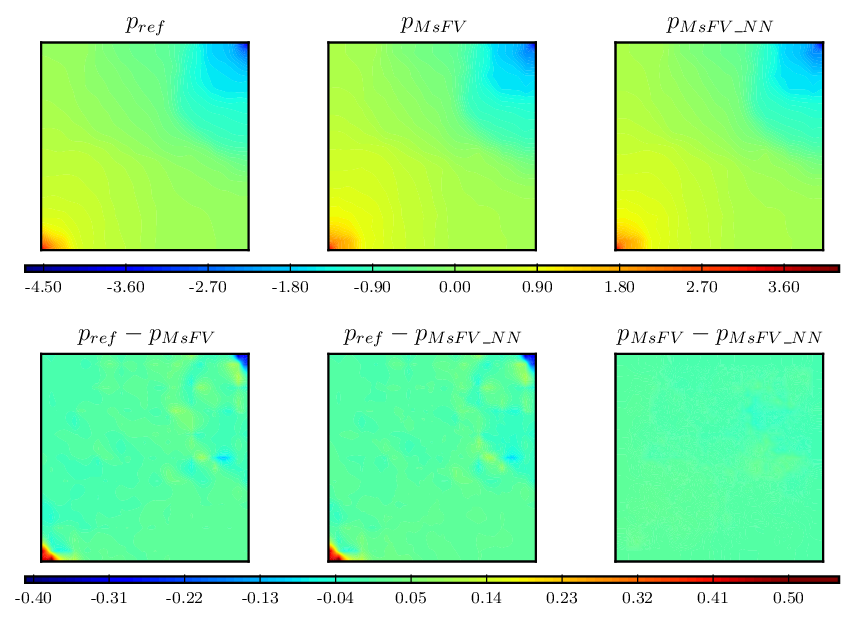}
    \caption{Pressure solution for one realization.}
  \end{subfigure}
  \begin{subfigure}{\linewidth}
  \centering
    \includegraphics[height=.45\textheight]{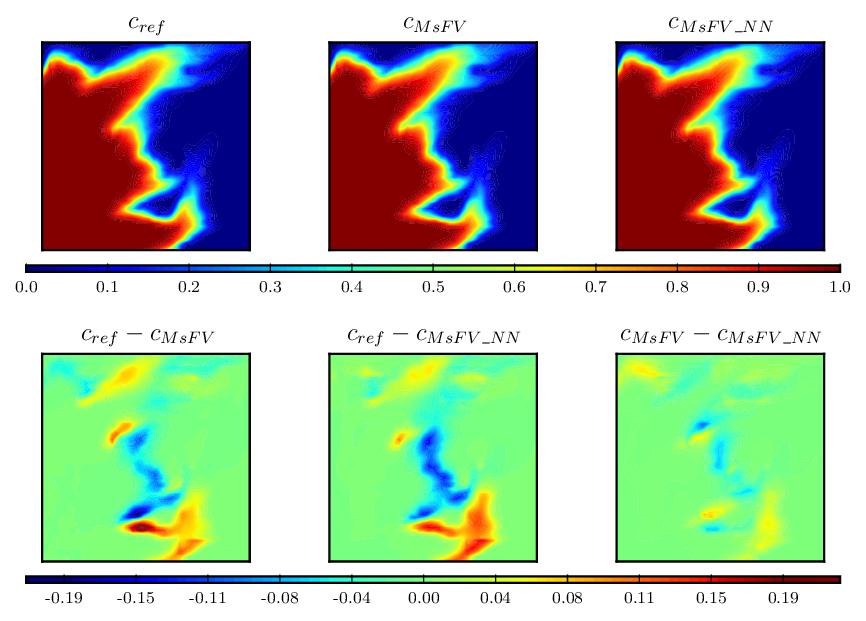}
    \caption{Concentration solution at $t=0.5$ PVI for one realization.}
  \end{subfigure}
  \caption{\textbf{Quarter-five spot problem:} Sample solution for one realization
    based on the reference (standard FVM), MsFV and MsFV-NN.}
  \label{fig:quarter_five_problem}
\end{figure}

\begin{figure}
  \centering
  \begin{subfigure}{\linewidth}
  \centering
    \includegraphics[height=.45\textheight]{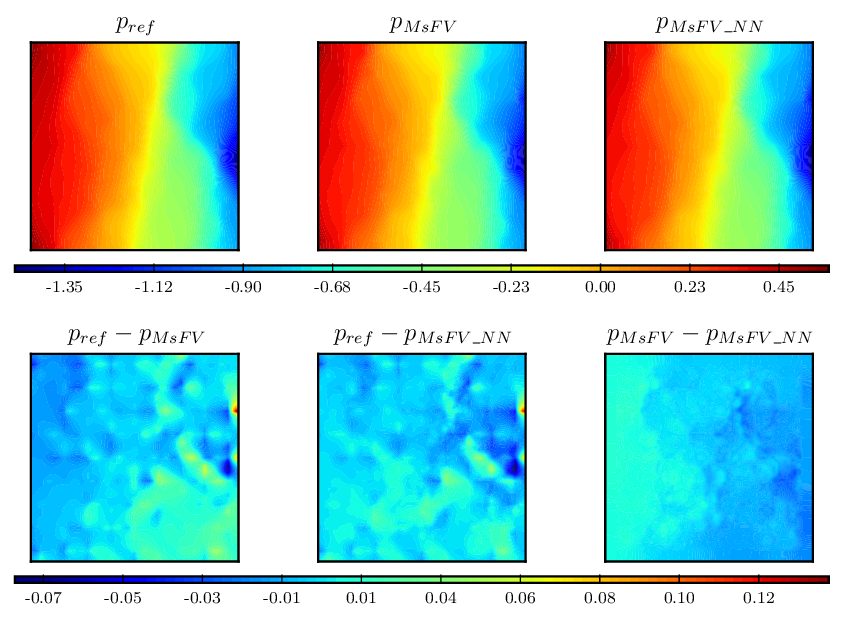}
    \caption{Pressure solution for one realization.}
  \end{subfigure}
  \begin{subfigure}{\linewidth}
  \centering
    \includegraphics[height=.45\textheight]{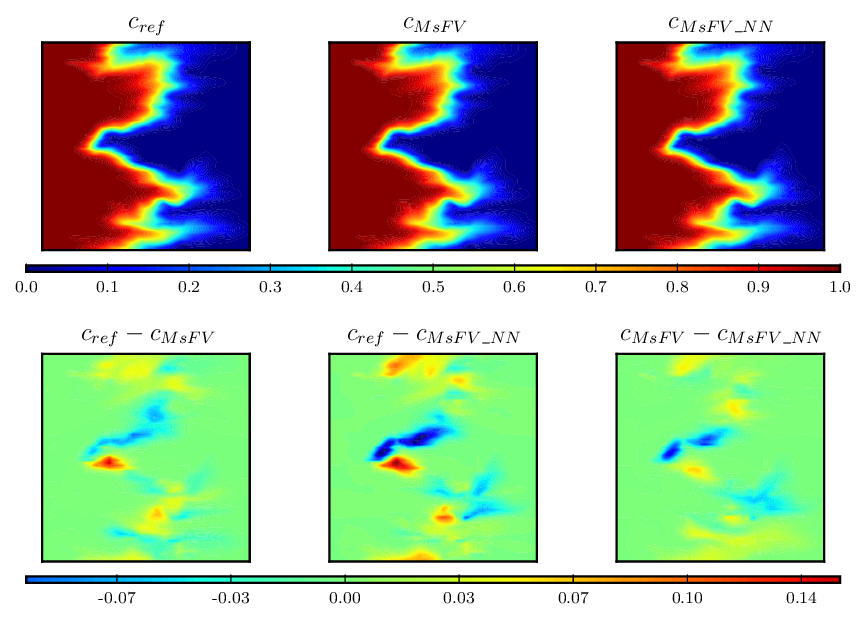}
    \caption{Concentration solution at $t=0.5$ PVI for one realization.}
  \end{subfigure}
  \caption{\textbf{Uniform flow problem:} Sample solution for one realization
    based on the reference (standard FVM), MsFV and MsFV-NN.}
  \label{fig:uniform_flow_problem}
\end{figure}

Figures~\ref{fig:quarter_five_problem}
and~\ref{fig:uniform_flow_problem} show sample solutions for one
realization of correlation length $L=0.1$, for the two test cases
considered. We also show the contour plot of the difference between
the reference and MsFV, the reference and MsFV-NN, and MsFV and
MsFV-NN.

\subsubsection{Comparison of errors}
\begin{figure} \centering

  \begin{subfigure}{\textwidth} \centering
    \begin{subfigure}{.32\textwidth}
      \includegraphics[width=\textwidth]{{{0.1_quarter_five_pressure_errors}}}
    \end{subfigure}
    \begin{subfigure}{.32\textwidth}
      \includegraphics[width=\textwidth]{{{0.1_quarter_five_velocity_errors}}}
    \end{subfigure}
    \begin{subfigure}{.32\textwidth}
      \includegraphics[width=\textwidth]{{{0.1_quarter_five_saturation_errors}}}
    \end{subfigure}
    \caption{Case $L=0.1$}
  \end{subfigure}

  \begin{subfigure}{\textwidth} \centering
    \begin{subfigure}{.32\textwidth}
      \includegraphics[width=\textwidth]{{{0.2_quarter_five_pressure_errors}}}
    \end{subfigure}
    \begin{subfigure}{.32\textwidth}
      \includegraphics[width=\textwidth]{{{0.2_quarter_five_velocity_errors}}}
    \end{subfigure}
    \begin{subfigure}{.32\textwidth}
      \includegraphics[width=\textwidth]{{{0.2_quarter_five_saturation_errors}}}
    \end{subfigure}
    \caption{Case $L=0.2$}
  \end{subfigure}

  \begin{subfigure}{\textwidth} \centering
    \begin{subfigure}{.32\textwidth}
      \includegraphics[width=\textwidth]{{{0.4_quarter_five_pressure_errors}}}
    \end{subfigure}
    \begin{subfigure}{.32\textwidth}
      \includegraphics[width=\textwidth]{{{0.4_quarter_five_velocity_errors}}}
    \end{subfigure}
    \begin{subfigure}{.32\textwidth}
      \includegraphics[width=\textwidth]{{{0.4_quarter_five_saturation_errors}}}
    \end{subfigure}
    \caption{Case $L=0.4$}
  \end{subfigure}

  \caption{\textbf{Quarter five spot problem:} Comparison of errors
    in MsFV and MsFV-NN.}
  \label{fig:quarter_five_errors}
\end{figure}
\begin{figure} \centering

  \begin{subfigure}{\textwidth} \centering
    \begin{subfigure}{.32\textwidth}
      \includegraphics[width=\textwidth]{{{0.1_uniform_flow_pressure_errors}}}
    \end{subfigure}
    \begin{subfigure}{.32\textwidth}
      \includegraphics[width=\textwidth]{{{0.1_uniform_flow_velocity_errors}}}
    \end{subfigure}
    \begin{subfigure}{.32\textwidth}
      \includegraphics[width=\textwidth]{{{0.1_uniform_flow_saturation_errors}}}
    \end{subfigure}
    \caption{Case $L=0.1$}
  \end{subfigure}

  \begin{subfigure}{\textwidth} \centering
    \begin{subfigure}{.32\textwidth}
      \includegraphics[width=\textwidth]{{{0.2_uniform_flow_pressure_errors}}}
    \end{subfigure}
    \begin{subfigure}{.32\textwidth}
      \includegraphics[width=\textwidth]{{{0.2_uniform_flow_velocity_errors}}}
    \end{subfigure}
    \begin{subfigure}{.32\textwidth}
      \includegraphics[width=\textwidth]{{{0.2_uniform_flow_saturation_errors}}}
    \end{subfigure}
    \caption{Case $L=0.2$}
  \end{subfigure}

  \begin{subfigure}{\textwidth} \centering
    \begin{subfigure}{.32\textwidth}
      \includegraphics[width=\textwidth]{{{0.4_uniform_flow_pressure_errors}}}
    \end{subfigure}
    \begin{subfigure}{.32\textwidth}
      \includegraphics[width=\textwidth]{{{0.4_uniform_flow_velocity_errors}}}
    \end{subfigure}
    \begin{subfigure}{.32\textwidth}
      \includegraphics[width=\textwidth]{{{0.4_uniform_flow_saturation_errors}}}
    \end{subfigure}
    \caption{Case $L=0.4$}
  \end{subfigure}

  \caption{\textbf{Uniform flow problem:} Comparison of errors
    in MsFV and MsFV-NN.}
  \label{fig:uniform_flow_errors}
\end{figure}

The errors of the solutions (pressure, velocity, concentration) of MsFV
and MsFV-NN are measured with respect to the reference solution using
an area weighted norm. Let $\mathbf u = (u_1, \cdots, u_n)$ be a vector of values
corresponding to cells $\Omega_1, \cdots, \Omega_n$ and let
$|\Omega_i|$ be the area of cell $i$, we define the area weighted
norm as $\lVert \mathbf u \rVert = (\sum_i {|u_i|^2|\Omega_i|})^{1/2}$. 
Using this notation, the pressure error ($e_p$), the
velocity error ($e_v$), and the concentration error ($e_c$) are:
\begin{align}
  e_p &= \frac{\lVert \mathbf p^{ref} - \mathbf p \rVert}{\lVert \mathbf p^{ref} \rVert} \\
  e_v &= \frac{\lVert \mathbf v_x^{ref} - \mathbf v_x \rVert}{\lVert \mathbf v_x^{ref} \rVert} + \frac{\lVert \mathbf v_y^{ref} - \mathbf v_y \rVert}{\lVert \mathbf v_y^{ref} \rVert}\\
  e_c &= \frac{1}{T}\int_0^T{\frac{\lVert \mathbf c^{ref}(\cdot,t) - \mathbf c(\cdot,t) \rVert}{\lVert \mathbf c^{ref}(\cdot,t)\rVert}dt}
\end{align}

Figures~\ref{fig:quarter_five_errors}
and~\ref{fig:uniform_flow_errors} show scatter plots of the errors
obtained by the MsFV and MsFV-NN. As expected, a better predictor
performance (in terms of the $R^2$-score) corresponded to a better
correlation between both errors.

\subsubsection{Estimated distributions}
\begin{figure} \centering

  \begin{subfigure}{\textwidth} \centering
    \begin{subfigure}{.32\textwidth}
      \includegraphics[width=\textwidth]{{{0.1_quarter_five_pressures_at_point}}}
    \end{subfigure}
    \begin{subfigure}{.32\textwidth}
      \includegraphics[width=\textwidth]{{{0.1_quarter_five_total_productions}}}
    \end{subfigure}
    \begin{subfigure}{.32\textwidth}
      \includegraphics[width=\textwidth]{{{0.1_quarter_five_wb_times}}}
    \end{subfigure}
    \caption{Case $L=0.1$}
  \end{subfigure}

  \begin{subfigure}{\textwidth} \centering
    \begin{subfigure}{.32\textwidth}
      \includegraphics[width=\textwidth]{{{0.2_quarter_five_pressures_at_point}}}
    \end{subfigure}
    \begin{subfigure}{.32\textwidth}
      \includegraphics[width=\textwidth]{{{0.2_quarter_five_total_productions}}}
    \end{subfigure}
    \begin{subfigure}{.32\textwidth}
      \includegraphics[width=\textwidth]{{{0.2_quarter_five_wb_times}}}
    \end{subfigure}
    \caption{Case $L=0.2$}
  \end{subfigure}

  \begin{subfigure}{\textwidth} \centering
    \begin{subfigure}{.32\textwidth}
      \includegraphics[width=\textwidth]{{{0.4_quarter_five_pressures_at_point}}}
    \end{subfigure}
    \begin{subfigure}{.32\textwidth}
      \includegraphics[width=\textwidth]{{{0.4_quarter_five_total_productions}}}
    \end{subfigure}
    \begin{subfigure}{.32\textwidth}
      \includegraphics[width=\textwidth]{{{0.4_quarter_five_wb_times}}}
    \end{subfigure}
    \caption{Case $L=0.4$}
  \end{subfigure}

  \caption{\textbf{Quarter five spot problem:} Estimated distributions
    by MsFV (orange dashed line), MsFV-NN (green dotted line), and
    reference (blue line).}
  \label{fig:quarter_five_uq}
\end{figure}
\begin{figure} \centering

  \begin{subfigure}{\textwidth} \centering
    \begin{subfigure}{.32\textwidth}
      \includegraphics[width=\textwidth]{{{0.1_uniform_flow_pressures_at_point}}}
    \end{subfigure}
    \begin{subfigure}{.32\textwidth}
      \includegraphics[width=\textwidth]{{{0.1_uniform_flow_total_productions}}}
    \end{subfigure}
    \begin{subfigure}{.32\textwidth}
      \includegraphics[width=\textwidth]{{{0.1_uniform_flow_wb_times}}}
    \end{subfigure}
    \caption{Case $L=0.1$}
  \end{subfigure}

  \begin{subfigure}{\textwidth} \centering
    \begin{subfigure}{.32\textwidth}
      \includegraphics[width=\textwidth]{{{0.2_uniform_flow_pressures_at_point}}}
    \end{subfigure}
    \begin{subfigure}{.32\textwidth}
      \includegraphics[width=\textwidth]{{{0.2_uniform_flow_total_productions}}}
    \end{subfigure}
    \begin{subfigure}{.32\textwidth}
      \includegraphics[width=\textwidth]{{{0.2_uniform_flow_wb_times}}}
    \end{subfigure}
    \caption{Case $L=0.2$}
  \end{subfigure}

  \begin{subfigure}{\textwidth} \centering
    \begin{subfigure}{.32\textwidth}
      \includegraphics[width=\textwidth]{{{0.4_uniform_flow_pressures_at_point}}}
    \end{subfigure}
    \begin{subfigure}{.32\textwidth}
      \includegraphics[width=\textwidth]{{{0.4_uniform_flow_total_productions}}}
    \end{subfigure}
    \begin{subfigure}{.32\textwidth}
      \includegraphics[width=\textwidth]{{{0.4_uniform_flow_wb_times}}}
    \end{subfigure}
    \caption{Case $L=0.4$}
  \end{subfigure}

  \caption{\textbf{Uniform flow problem:} Estimated distributions
    by MsFV (orange dashed line), MsFV-NN (green dotted line), and
    reference (blue line).}
  \label{fig:uniform_flow_uq}
\end{figure}

Finally, we compare all three methods in an uncertainty quantification
task where we estimate the pressure $p$ at $(1/4,1/4)$, the total
production $Q$, and the water breakthrough time $t_{wb}$ (time when water
fraction reaches 1\% at the production well). 

Figures~\ref{fig:quarter_five_uq}
and~\ref{fig:uniform_flow_uq} show the estimated distributions
according to each method. We can see that the distributions given by
MsFV and MsFV-NN are almost indistinguishable even for the less
accurate predictor ($L=0.1$). From these results, it is clear that the
effectiveness of the hybrid model is attached to the effectiveness of
the target model (MsFV). The hybrid model is expected to perform well
as long as the target model itself serves as a good proxy to the
``true'' solution.

\subsection{Hyperparameter tuning} \label{sec:hyperopt}
\begin{table}
  \centering
  \caption{Results of hyperparameter tuning.}
  \begin{tabular}{lr}
    \hline
    Dropout rate & $5.4\%$ \\
    Learning rate & $3.1\times 10^{-3}$ \\
    $R^2$-score & $0.97$ \\
    \hline
  \end{tabular}
  \label{table:netopt}
\end{table}
In this section, we show how to further improve the learning
performance by fine-tuning the model with a hyperparameter
optimization algorithm. Specifically, we employ the Tree-Parzen
Estimator algorithm. We shall consider the case of $L=0.1$ where the
trained predictor performed with a score of $0.927$.

Previously, a dropout rate of $5\%$ and a default learning rate of
$10^{-3}$ have been fixed. Here we let the hyperparameter optimization algorithm
tune the values for the dropout rate and the learning rate. Moreover,
we also employ batch normalization to enhance the optimization process.

Table~\ref{table:netopt} summarizes the
hyperparameter optimization results under a budget of $20$
iterations where we observe significant improvement in the
$R^2$-score. Figure~\ref{fig:errors_tuned} shows the error scatter
plot for the resulting hybrid model. An improvement in the correlation
is observed (please compare with Figures~\ref{fig:quarter_five_errors}
and~\ref{fig:uniform_flow_errors}). Figure~\ref{fig:uq_tuned} compares
the estimated distributions for the quantities of interest where a
strong agreement between the data-driven approach and the MsFV is observed. Finally,
Table~\ref{table:summary_tuned} presents summary statistics of the
results obtained.
Overall, we see that there are improvements in both the errors and the
estimated distributions when the learning performance increases, as is
expected. Of course, even more improvements can be achieved by
further tuning the model, for example by increasing the number of iterations
of the hyperparameter optimization, or by employing additional tools such
as $L^1$ and $L^2$ regularizers.

\begin{figure}
  \centering
  \begin{subfigure}{\textwidth}
    \begin{subfigure}{.32\textwidth}
      \includegraphics[width=\textwidth]{{{0.1_tuned_quarter_five_pressure_errors}}}
    \end{subfigure}
    \begin{subfigure}{.32\textwidth}
      \includegraphics[width=\textwidth]{{{0.1_tuned_quarter_five_velocity_errors}}}
    \end{subfigure}
    \begin{subfigure}{.32\textwidth}
      \includegraphics[width=\textwidth]{{{0.1_tuned_quarter_five_saturation_errors}}}
    \end{subfigure}
    \caption{Quarter five spot problem ($L=0.1$)}
  \end{subfigure}

  \begin{subfigure}{\textwidth}
    \begin{subfigure}{.32\textwidth}
      \includegraphics[width=\textwidth]{{{0.1_tuned_uniform_flow_pressure_errors}}}
    \end{subfigure}
    \begin{subfigure}{.32\textwidth}
      \includegraphics[width=\textwidth]{{{0.1_tuned_uniform_flow_velocity_errors}}}
    \end{subfigure}
    \begin{subfigure}{.32\textwidth}
      \includegraphics[width=\textwidth]{{{0.1_tuned_uniform_flow_saturation_errors}}}
    \end{subfigure}
    \caption{Uniform flow problem ($L=0.1$)}
  \end{subfigure}
  \caption{Comparison of errors in MsFV and MsFV-NN (tuned model,
    $L=0.1$). Improvements can be seen with respect to
    Figures~\ref{fig:quarter_five_errors}(a)
    and~\ref{fig:uniform_flow_errors}(a).}
  \label{fig:errors_tuned}
\end{figure}

\begin{figure}
  \centering
  \begin{subfigure}{\textwidth}
    \begin{subfigure}{.32\textwidth}
      \includegraphics[width=\textwidth]{{{0.1_tuned_quarter_five_pressures_at_point}}}
    \end{subfigure}
    \begin{subfigure}{.32\textwidth}
      \includegraphics[width=\textwidth]{{{0.1_tuned_quarter_five_total_productions}}}
    \end{subfigure}
    \begin{subfigure}{.32\textwidth}
      \includegraphics[width=\textwidth]{{{0.1_tuned_quarter_five_wb_times}}}
    \end{subfigure}
    \caption{Quarter five spot problem ($L=0.1$)}
  \end{subfigure}
  \begin{subfigure}{\textwidth}
    \begin{subfigure}{.32\textwidth}
      \includegraphics[width=\textwidth]{{{0.1_tuned_uniform_flow_pressures_at_point}}}
    \end{subfigure}
    \begin{subfigure}{.32\textwidth}
      \includegraphics[width=\textwidth]{{{0.1_tuned_uniform_flow_total_productions}}}
    \end{subfigure}
    \begin{subfigure}{.32\textwidth}
      \includegraphics[width=\textwidth]{{{0.1_tuned_uniform_flow_wb_times}}}
    \end{subfigure}
    \caption{Uniform flow problem ($L=0.1$)}
  \end{subfigure}
  \caption{Estimated distributions (tuned model, $L=0.1$)
    by MsFV (orange dashed line), MsFV-NN (green dotted line), and
    reference (blue line).
    Compare with Figures~\ref{fig:quarter_five_uq}(a)
    and~\ref{fig:uniform_flow_uq}(a).}
  \label{fig:uq_tuned}
\end{figure}

\begin{table}
  \centering
  \caption{Summary statistics and point estimates ($L=0.1$).}
  \begin{subtable}{\textwidth} \centering
    \caption{Quarter five spot problem}
    \begin{tabular}{lrrrr}
      \hline
                 & Reference & MsFV & MsFV-NN & Untuned \\
      \hline
      $\bar e_p$                     & -      & 0.0525 & 0.0560 & 0.0586 \\
      $s_{e_p}$            & -      & 0.0228 & 0.0226 & 0.0232 \\
      $\bar e_v$                     & -      & 0.1312 & 0.1463 & 0.1654 \\
      $s_{e_v}$            & -      & 0.0216 & 0.0243 & 0.0231 \\
      $\bar e_c$                     & -      & 0.0268 & 0.0298 & 0.0326 \\
      $s_{e_c}$           & -      & 0.0053 & 0.0068 & 0.0058 \\
      $\bar{p}_{(1/4,1/4)}$           & 0.5283 & 0.5297 & 0.5303 & 0.5240 \\
      $s_{p_{(1/4,1/4)}}$   & 0.2075 & 0.2081 & 0.2092 & 0.2036 \\
      $\bar{Q}$                      & 0.1910 & 0.1906 & 0.1906 & 0.1904 \\
      $s_Q$               & 0.0060 & 0.0062 & 0.0063 & 0.0062 \\
      $\bar{t}_{wb}$                 & 0.4488 & 0.4404 & 0.4406 & 0.4402 \\
      $s_{t_{wb}}$        & 0.0639 & 0.0646 & 0.0650 & 0.0639 \\
      \hline
    \end{tabular}
    \label{table:quarter_five_tuned_summary}
  \end{subtable}

  \vspace{1em}
  \begin{subtable}{\textwidth}
    \centering
    \caption{Uniform flow problem}

\begin{tabular}{lrrrr}
  \hline
  & Reference & MsFV & MsFV-NN & Untuned \\
  \hline
  $\bar e_p$                       & -      & 0.0301 &  0.0357 & 0.0408 \\
  $s_{e_p}$             & -      & 0.0093 &  0.0105 & 0.0141 \\
  $\bar e_v$                       & -      & 0.2412 &  0.2876 & 0.3504 \\
  $s_{e_v}$             & -      & 0.0267 &  0.0338 & 0.0420 \\
  $\bar e_c$                       & -      & 0.0205 &  0.0228 & 0.0267 \\
  $s_{e_c}$             & -      & 0.0039 &  0.0043 & 0.0050 \\
  $\bar{p}_{(1/4,1/4)}$             & 0.2724 & 0.2722 &  0.2732 & 0.2677 \\
  $s_{p_{(1/4,1/4)}}$    & 0.1160 & 0.1162 &  0.1181 & 0.1135 \\
  $\bar{Q}$                        & 0.2339 & 0.2345 &  0.2352 & 0.2365 \\
  $s_Q$                 & 0.0589 & 0.059  &  0.0590 & 0.0591 \\
  $\bar{t}_{wb}$                   & 0.7460 & 0.7423 &  0.7436 & 0.7479 \\
  $s_{t_{wb}}$          & 0.1746 & 0.1743 &  0.1738 & 0.1737 \\
  \hline
\end{tabular}
    
    \label{table:uniform_flow_tuned_summary}
  \end{subtable}
  \label{table:summary_tuned}
\end{table}

\subsection{Computational gains}
\begin{table} \centering
\caption{Time to generate $1000$ basis functions using different methods.}
\begin{tabular}{lr}
  \hline
  Method & Time [sec]\\
  \hline
   Sparse direct solver (UMFPACK) & 2.14 \\ 
  GMRES (\texttt{tol=1e-5}) & 7.21 \\ 
  GMRES (\texttt{tol=1e-16}) & 23.77 \\ 
  NN prediction & 0.389 \\
  NN (batch eval) & 0.083 \\
  NN (batch eval) (GPU) & 0.017 \\
  \hline
\end{tabular}
\label{table:times}
\end{table}

For an estimate of the computational speedup provided by the proposed
method, we compared the time taken to generate $1000$ basis functions using
the predictor vs. the standard approach of solving local problems.
Since the MsFV method obtains the basis functions by solving local
problems which involve many intermediary steps, and this could lead to
data-derived overheads which are implementation-dependent, we
decided to measure the time of
solving the four local \emph{2D problems only}, i.e. without
accounting for the overheads of getting the local boundary conditions
(which are obtained by solving the 1D problems) and assembling the
local matrices. We employed two solvers for the local problems: GMRES
iterative solver and UMFPACK direct solver, both highly optimized
C-compiled packages provided in \texttt{numpy}/\texttt{scipy}.

Table~\ref{table:times} summarises the run times obtained. These
results were obtained using one thread (except for the last row which
is run on GPU). Here, ``batch eval'' refers to the prediction of the
$N$ basis functions ``at once'': for a given input vector
$\kappa_i, i=1,2,...,N$, the predictor performs a matrix-vector
multiplication on $\kappa_i$. But this can be implemented as a
matrix-matrix multiplication simply by building the matrix
$\mathcal{K}$ whose columns are the vectors $\kappa_i$, allowing for
additional numerical optimizations.

In this case, we see that the direct solver outperformed the iterative
solver for the local problems since the local matrices are small,
which is the common scenario in multiscale methods. We also see that
the data-driven approach clearly outperforms the solver component, and if we
add the overheads of solving the 1D problems plus the local matrix
assembly, the computational advantage will be amplified. We note
however that these times can vary depending on the implementation. In
particular, different neural network architectures and different
solvers for the system of equations may yield different
times. Nevertheless, it is unlikely that solving the local problems
will outperform a forward pass of a neural network, i.e. direct
matrix-vector computations.

\section{Conclusions and remarks}
We have seen that for the presented subsurface flow problems, shallow neural networks
performed very well as a simple
surrogate for the computation of basis functions in the multiscale
finite volume method. Further, we 
draw the following remarks:
\begin{itemize}
\item[--] Results obtained for uncertainty propagation using MsFV and
  the proposed MsFV-NN method were practically indistinguishable.
\item[--] The proposed method is applicable to any multiscale method
  where the sub-grid scales are captured numerically by solving local problems.
\item[--] The proposed method is scalable with large coarse partitions
  (since more data samples are obtained per simulation run).
\end{itemize}

In addition, we note that if the \emph{data distribution} remains
unchanged (or is similar to that of the training data), then the same
trained predictor can be used for different problem conditions (for
example, to perform well location optimization), and further
computational gains can be achieved since we avoid training a new
predictor. This is the situation in cases such as steady state flow or
tracer flow.

We have presented the first application of machine learning to
capture sub-grid scale heterogeneities within a multiscale method.  As
our next step, we aim to study the application of the presented method
for multiphase flow in porous media. Other possible research
directions include extensions to more general permeability fields
with anisotropy and channelized structures.

\bibliographystyle{plainnat}
\bibliography{biblio}
\end{document}